\documentclass{article}
\usepackage{arxiv}

\usepackage{graphicx}
\usepackage{subfig}
\usepackage{amsmath}

\usepackage{epsfig} 
\usepackage{mathptmx} 
\usepackage{times} 
\usepackage{amssymb}  
\usepackage{booktabs}
\usepackage{multirow}
\usepackage{algpseudocode}
\usepackage{float}
\usepackage[linesnumbered,noend,ruled]{algorithm2e}
\usepackage{graphicx}
\usepackage{textcomp}
\usepackage{xcolor}
\usepackage{stackengine}

\makeatletter
\renewcommand{\@algocf@capt@plain}{above}
\makeatother

\IncMargin{1em}
\usepackage{hyperref}

\usepackage{comment}
\usepackage{booktabs}
\usepackage{multirow}
\usepackage{algpseudocode}
\usepackage{float}
\usepackage{graphicx}
\usepackage{textcomp}
\usepackage{xcolor}
\usepackage{stackengine}
\usepackage{pgfplots}
\usepackage{pgfplotstable}
\usepackage{ifthen}
\pgfplotsset{compat=1.7}
\usepgfplotslibrary{groupplots}
\usepackage{verbatim}
\usepackage{balance}
\usepackage{apalike}

\usepackage{pifont}

\title{The BS-meter: Detecting Politics and Labour through ChatGPT’s Language}

\author{
 Alessandro Trevisan \\
  University of Cambridge\\
  \texttt{at2005@cam.ac.uk} \\
   \And
 Harry Giddens \\
  University of Cambridge\\
  \texttt{harygiddens@gmail.com} \\
  \And
 Sarah Dillon \\
  University of Cambridge\\
  \texttt{sjd27@cam.ac.uk} \\
  \And
 Alan F. Blackwell \\
  University of Cambridge\\
  \texttt{afb21@cam.ac.uk}
}

\begin{document}

\maketitle
\thispagestyle{empty}
\pagestyle{empty}

\begin{abstract}
What can we learn about language from studying how it is used by ChatGPT and other large language model (LLM)-based chatbots? In this paper, we analyse the distinctive character of language generated by ChatGPT, in relation to questions raised by natural language processing pioneer, and student of Wittgenstein, Margaret Masterman. Following frequent complaints that LLM-based chatbots produce ``bullshit,'' in the sense of Frankfurt’s popular monograph \textit{On Bullshit}, we conduct an empirical study to contrast the language of 1,000 scientific publications with typical text generated by ChatGPT. We then explore whether the same language features can be detected in two well-known contexts of social dysfunction: George Orwell’s critique of political speech, and David Graeber’s characterisation of bullshit jobs. Using simple hypothesis-testing methods, we demonstrate that a statistical model of bullshit can reliably relate the Frankfurtian artificial bullshit of ChatGPT to the political and workplace functions of bullshit as observed in natural human language.
\end{abstract}

\section{Introduction}
Margaret Masterman, founder of the Cambridge Language Research Unit (CLRU) and student of Wittgenstein, was an AI pioneer whose work on the philosophy of language and machine translation is newly relevant to Large Language Models (LLMs). In this paper, we report an empirical investigation of LLM behaviour that offers wider insights into language use, in addition to clarifying the social and epistemological status of LLMs themselves. We draw on two specific aspects of Masterman’s work. The first is her ambition (as she said, not realisable with the technology of her time) to create a \textit{semantic detector},\footnote{``1. The Need for the Establishment of a Semantic Discipline of `Message-Detection' for Machine Translation / l.i. Comment on the Theoretically Unsatisfactory Nature of the Present Situation / [...The] mechanizable techniques at present being used to analyse language are not powerful enough to detect the message, or argument, of any particular text. [...] What is needed is a discipline which will study semantic message-connection.'' \cite[p. 438]{masterman1961semantic}} able to identify the actual meaning of a message, rather than simply translate the words being used \cite{masterman1961semantic}. The second is the way her investigations of embodied aspects of language extended beyond mere text, encompassing phenomena such as breath, phrasing and the contexts of speech.

The technical work reported in the second part of the paper describes the implementation of a Masterman Semantic Detector (MSD),\footnote{This is our coinage, proposed in honour of Masterman, rather than her own proposal.} developed out of concern with the kinds of LLM-generated text that are frequently described by commentators and researchers as ``slop'' \cite{adami_ai-generated_2024,copestake2024llms,gioia_new_2024,malik_ai_2025} or ``bullshit'' \cite{bernoff_chatgpt_2022,vincent_ai-generated_2022,katwala_chatgpts_nodate,narayanan_chatgpt_2022,deck2023bullshit,blackwell_oops_2023,gershon2023bullshit,sundar2023calling,hicks2024chatgpt,hannigan2024beware,vallor2024ai,haigh_artificial_2025}. Such works do not use the term ``bullshit'' colloquially but technically, often drawing on a core idea from Harry Frankfurt’s \cite{frankfurt2005bullshit} philosophical account of the term: namely, they understand bullshit as a form of linguistic communication characterised by a ``lack of connection to a concern with truth – […] indifference to how things really are'' \cite[pp. 33-34]{frankfurt2005bullshit}. LLMs, indeed, are trained to predict the most likely next “token”—typically a word, a sub-word unit, or other linguistic symbol—in a textual sequence, not to directly encode or manipulate facts.

The theoretical work in the first part of the paper explains how this kind of language arises, in relation to the metaphorical ``body'' of LLM-based chatbots, through the software architectures and business models that might be considered as the animating breath, algorithmic phrasing and social contexts of the fictional personas now being marketed as ``Gemini,'' ``Claude,'' ``ChatGPT,'' and so on. In the second part of the paper, we investigate empirically whether this kind of BS/sloppy semantic content, newly seen as undesirable in mechanically synthesised text, was already present in human writing. The result demonstrates an apparently effective ``BS-meter.''



\maketitle

\section{The linguistic ``body'' of ChatGPT}

Does ChatGPT have a body? ChatGPT itself replies: ``As a large language model, I don't have a physical body.'' Yet Critical AI scholars are aware, as brilliantly illustrated in Crawford and Joler's \cite{crawford2018anatomy} ``Anatomy of an AI System,'' that these systems are only too physical, composed of ghost labour, surveillance capitalism and wholesale theft of copyright, not to mention the environmental and neo-colonial harms of the whole infrastructure that accompanies the business models of AI and social media \cite{couldry2020costs,crawford2021atlas,muldoon2024feeding}. We relate these complexities to the work of Masterman, one of the students whom Wittgenstein entrusted with compiling his lecture notes into what became \textit{The Blue Book}. Masterman campaigned throughout her career to probe beyond the reductive formal accounts of language that she found inadequate in the computational approaches advocated by Chomsky, Turing, and other AI pioneers \cite{liu2021wittgenstein}.\footnote{Lydia Liu \cite[p. 450]{liu2021wittgenstein} suggests that ``AI scientists today would be inclined to translate [Masterman’s] term `Roget’s unconscious' into neural networks,'' although Masterman’s colleague Yorick Wilks, in his editorial introduction to her collected works, felt she would have been unimpressed by the thinness of the statistical ``connectionism'' of neural network machine learning \cite[p. 16]{masterman2005language}.}

Masterman’s technical work in the 1950s and `60s, especially on the conceptual networks expressed in thesauri, anticipated the corpus-driven computational methods that paved the way for LLMs, including vector semantics, word embeddings, and the TF*IDF measure attributed to Masterman’s student Karen Sp{\"a}rck Jones \cite{sparck1972statistical}. Over 60 years ago, it was impossible for Masterman to imagine the scale of textual corpora that can now be extracted from billions of Internet users, or the network algorithms designed for computers a billion times more powerful than those used at CLRU. Although Masterman did not anticipate this kind of non-human ``body'' in her writing, and we have only her views on those aspects of the human body that she understood as challenges to reductive models of text,\footnote{The influence of Wittgenstein’s thought on Masterman’s consideration of the embodied foundations of language is evident, for example, in her investigations of phrasing \cite[p. 253ff]{masterman2005language} and of context (p. 124). Future work might profitably consider what the metaphorical correspondences of breath, phrasing and bodily context might be for an entity such as ChatGPT.} the behaviour emerging from this massive complexity can be understood in terms of the Wittgensteinian concepts \textit{Lebensform} and \textit{language-game}. \textit{Lebensform} is the social context.\footnote{For a detailed interpretation of \textit{Lebensform}, see \cite{gier_wittgenstein_1980}.} \textit{Language-game} is Wittgenstein’s understanding that language can only make sense in that social context.\footnote{For a more specific definition of what a language-game is, consider Wittgenstein’s own introduction of the term in the \textit{Philosophical Investigations} \cite[§7]{wittgenstein_philosophical_2009}: ``I shall also call the whole, consisting of language and the activities into which it is woven, a `language-game'.'' Wittgenstein himself then offers some examples of language-games (§23): ``Giving orders, and acting on them — […] Reporting an event — […] Guessing riddles — Cracking a joke; telling one — […] Requesting, thanking, cursing, greeting, praying.''} 

Applying this conception of language to the outputs of LLM-based chatbots, we ask: what \textit{Lebensform} does this language belong to? This calls for investigation of the forces at play in moulding an LLM-based chatbot’s language: from its technical components, to the datasets it was trained on, to the communities of annotators employed to optimise its responses. We are inspired by Gier’s \cite{gier_wittgenstein_1980} layered interpretation of \textit{Lebensform} to conceptualise LLM-based chatbots as socio-linguistic-technical assemblages whose component parts are relevant to understanding their language.

\subsection{Dissecting LLMs}

The analogy between human bodies and computing infrastructure is nearly 200 years old. Simon Schaffer \cite[p. 210]{schaffer1994babbage} reports how Charles Babbage's promotion of his computing machinery as a manufactory for numbers ``precisely embodied the intelligence of theory and abrogated the individual intelligence of the worker.'' Anticipating today's AI entrepreneurs, ``[t]he faculties of memory and foresight with which Babbage sought to endow the Analytical Engine also characterize his self-presentation as the unique author of the machine. They embodied his control over the engine while they disembodied the skills and camouflaged the work force on which it depended'' \cite[p. 214]{schaffer1994babbage}. Schaffer traces Babbage’s influence on the now-familiar observation by Karl Marx \cite{marx_grundrisse_1973} that the factory system is a ``mighty organism'' where ``it is the machine which possesses skill and strength in place of the worker''; we bring this logic to our own study of the language-using anatomy of ChatGPT.

Today’s AI promoters have invested so much effort to present their systems as autonomously intelligent that care is needed to disentangle two things: the actual human language that has been encoded in the training data of LLMs (including many kinds of language-game), in contrast to the novel ways that language might be used by the ``mighty organism'' of ChatGPT – a new kind of socio-linguistic-technical system that we interpret in relation to Wittgenstein’s \textit{Lebensform}. 

We emphasise the key anatomical distinction between, on one hand, the generalised \textit{sequence-prediction capabilities} of LLMs that have been trained with samples of natural human language, and on the other, the way that these capabilities are used to construct \textit{LLM-based chatbots} \cite{stone2024origins}. In one famous lineage, the first three versions of OpenAI’s original GPT (Generative Pretrained Transformer) LLM attracted little public attention,\footnote{Note that RoBERTa \cite{liu2019roberta}, which we used in the MSD experiments reported below, is another early language model founded on the transformer architecture developed at Google \cite{vaswani_attention_2017}, and that it also precedes ChatGPT \cite{OpenAI_introducing_2022} and the innovation of the LLM-based chatbot.} while the LLM-based chatbot ChatGPT was a blockbuster success. The technical components needed to turn an LLM into an LLM-based chatbot are described by Shanahan \cite[p. 74]{shanahan2024talking} as a supplementary \textit{dialogue management system} (DMS). A product such as ChatGPT consists of two elements: first, the underlying pretrained LLM, and second, the supplementary DMS.

\subsection{The Anatomy of Bullshit}

Because the first component – the pretrained LLM – encodes a huge corpus of actual human language use, of language-games, it can be a valuable research instrument. In addition to the ``games'' played in everyday language such as ``giving orders, and acting on them,'' ``reporting an event,'' ``cracking a joke,'' ``requesting, thanking, cursing, greeting, praying,'' etc. \cite[§23]{wittgenstein_philosophical_2009}, LLM training data can be used to study more formal situations. For example, cognitive scientist Clayton Lewis \cite{lewis2025artificial} uses ChatGPT as an instrument for ``Artificial Psychology,'' revisiting the controlled language of psychological experiments, such as logical reasoning, analogy, or short-term memory. Our investigation of bullshit relies on the fact that the LLM training data includes these language-games, along with Wittgenstein’s social exchanges, Lewis’s cognitive tasks, and many others. 

We also understand this type of language as being instilled in the model by diverse socio-cultural and ideological currents, active in the environments in which it was designed and deployed. ChatGPT’s acquisition of bullshit can also be understood as the product of processes of work automation and, more broadly, of late capitalism. As identified by Marx and, more recently, Pasquinelli \cite{pasquinelli2023eye}, ever since the Industrial Revolution, labour has increasingly been shaped to fit within rigid processes of automated production.\footnote{We also note David Runciman’s observation that artificial intelligence, in this systemic sense, has existed ever since corporations were defined as artificial legal persons \cite{azhar_superintelligence_2020}.} This labour milieu has led to the production of machines (LLM-based chatbots) that mimic these dynamics in their language, that speak in this mechanically-embodied fashion.

We contend that the general-purpose capabilities of the pretrained LLMs, which encode \textit{many} naturally occurring kinds of language-game, from logic and humour to bullshit and slop, are in practice filtered by the second component – the DMS – to emphasise the latter rather than the former. In other words, these chatbots are designed to present themselves as intelligent and reliable assistants, even though their underlying mechanism is a probabilistic system trained to predict patterns in language rather than to distinguish truth from falsehood. Given the huge commercial value of the transition from LLM to LLM-based chatbot, it is unsurprising that the details of the DMS are highly secret, often not even mentioned, to an extent that many commentators speak of the LLM as if it were the whole product. There are few published descriptions of the DMS component, and statements made in public are not necessarily to be trusted, given the billion-dollar investments depending on their reception and the potentially unethical practices employed \cite{muldoon2024feeding}. However, it is reasonably certain that these modules (sometimes called ``guardrails'' when a company wishes to emphasise their concern for consumer safety) employ methods such as instruction-tuning, prompt-engineering, and reinforcement learning from human feedback (RLHF) \cite{stone2024origins}.

\subsection{The Poetics of an LLM}

As an aside considering the other ways in which the underlying pretrained LLM \textit{might} be used, a disregard for truth and falsity does not automatically render an utterance bullshit. Consider, for instance, Aristotle’s \cite[9.1451a36–39]{aristotle-poetics} understanding that ``the job of the poet is not relating what actually happened, but rather the kind of thing that \textit{would} happen---that is to say, what is possible in terms of probability and necessity.'' Or consider Heidegger’s \cite[§35]{heidegger_being_time_1967} notion of \textit{idle talk}, a mode of discourse in which language circulates independently of genuine understanding and amounts to the “passing along” of what has already been said. Among other differences that may be pointed out between these linguistic practices, bullshit differs in that it implies a degree of deception on the part of the bullshitter. In distinguishing the liar from the bullshitter, Frankfurt \cite[p. 54]{frankfurt2005bullshit} notes that “[t]he bullshitter may not deceive us, or even intend to do so, either about the facts or about what he takes the facts to be. What he does necessarily attempt to deceive us about is his enterprise. His only indispensably distinctive characteristic is that in a certain way he misrepresents what he is up to.” We thus wish to draw attention to a notion often overlooked in the literature on LLMs and bullshit: namely, that LLM outputs would not inherently qualify as bullshit if they were presented for what they truly are---predicted continuation sequences. However, the companies that deploy these models often construct the DMS to mask this limitation, presenting the generated texts as if they were the product of an intelligent, knowledgeable assistant. The language-game of bullshit that results from the operation of the DMS is precisely this language of deception.

\subsection{The DMS as Paratext}

Shanahan et al. \cite{shanahan2024role} describe the distinction between the many \textit{potential} behaviours encoded in the underlying LLM and the \textit{actual} behaviour resulting from the DMS as ``role play.'' While Shanahan’s \cite{Shanahan_2010_Embodiment,shanahan2024talking} analysis also draws on Wittgenstein’s philosophy of language, our interpretation of the DMS is more literary. We suggest that the relationship between an LLM and an LLM-based chatbot can be understood in terms of the distinction in literary theory between text and paratext.\footnote{Genette’s concept of paratexuality is one of his five categories of transtextuality, the others being intertextuality, metatextuality, architextuality, and hypertextuality. It would be possible, and perhaps highly productive, to produce a Genettian reading of LLMs in general, for example, developing an account of an LLMs operations by analysing its architextuality, defined by Genette \cite[p. 1]{genette1997paratexts} as ``the entire set of general or transcendent categories---types of discourse, modes of enunciation, literary genres---from which emerges a singular text.'' But such an endeavour goes beyond our purpose for this paper, which is to focus on just one aspect---the DMS---and to offer an alternative, and we think more productive, metaphoric than those currently circulating regarding it.} The paratext includes all the material one finds \textit{around} the main text in the copy of a work, such as the name of the author, the title of the work, the subtitle, preface, publisher, the typesetting, cover image, blurb, endorsements and so on. It also extends beyond the physical form of the text, to the marketing, reviews, author interviews, private correspondence, diaries, and so on. The paratext is, as Lejeune \cite[p. 45]{philippe1975pacte} understands it, ``a fringe of the printed text which in reality controls one’s whole reading of the text'' (quoted in \cite[p.2]{genette1997paratexts}). Genette \cite[p. 2]{genette1997paratexts} describes this as a ``\textit{threshold}'' - ``a zone not only of transition but also of \textit{transaction}: a privileged space of a pragmatics and a strategy.'' He is generous in his interpretation of its purpose, seeing it as aimed ``at the service of a better reception for the text and a more pertinent reading of it'' \cite[p. 2]{genette1997paratexts}. But he does caveat that pertinence is always ``in the eyes of the author and his allies'' \cite[p. 2]{genette1997paratexts}. A paratext is therefore always designed to influence public reception of a text in line with how the author and their allies wish that text to be received. 

In our analysis, the DMS of ChatGPT – the hidden prompts, guardrails, tuning and so on – constitutes a paratext that is intended to present the statistical sequence-generating capabilities of the underlying LLM \textit{as if it were} a conversation between living persons, rather than a simple series of probabilistically-chosen words. The statistical tendencies of the base model continue to govern text production in an LLM-based chatbot, but they are articulated through the policies imposed by the DMS, and may in some cases be fully overridden when outputs are deemed biased or unsafe. This kind of paratext is therefore far more significant than the paratexts envisaged by Genette. Whereas Genette's paratexts merely accompany a text and shape its reception, the DMS does not simply accompany the statistical generations of an LLM; it actively conceals those ``raw'' generations behind a layer of imposed discourse.

Alternative paratexts could have determined the reception of ChatGPT’s outputs quite differently.\footnote{Think, for instance, about how different types of DMS engender the different GPTs now made available by OpenAI alongside their general purpose, flagship model: these are bots with different ``personalities'' (whose raw statistical predictions come ``dressed'' in different paratexts), designed to assist with individual specialised tasks, like cooking or researching academic literature.} Genette \cite[p. 2]{genette1997paratexts} gives the following example: ``limited to the text alone and without a guiding set of directions, how would we read Joyce’s \textit{Ulysses} if it were not entitled \textit{Ulysses}?'' Similarly, if, as Shanahan \cite[p. 70]{shanahan2024talking} suggests, every output of an LLM response was preceded by a phrase such as ``given the statistical distribution of words….,'' how would that determine the way users perceived and interacted with it, and how its author and their allies were able to present it? Or, again, how would we perceive an LLM-based chatbot’s outputs if we could see the hidden prompts it receives before addressing our own question?

The DMS-paratext designed for ChatGPT exploits the social cognition tendency of the human brain to anthropomorphise,\footnote{See Sundar and Liao's \cite{sundar2023calling} discussion of human psychology, human-computer interaction and the ``Computers are Social Actors'' research programme, and references therein, in particular the collaborative work of Clifford Nass.} and can be understood as part of the history of disingenuous rhetoric that continues to surround AI, influencing users’ perceptions of AI technologies.\footnote{As Syed Mustafa Ali et al. \cite[p. 1]{ali2023histories} note, ``the history of imaginative thinking around AI, in fact and fiction, influences how AI is produced, perceived and regulated, and the rhetorical framing of `AI’, past and present, by scientists, technologists, governments, corporations, activists and the media, performatively creates and shapes the very phenomenon purportedly under analysis.'' See also \cite{hunter1991rhetoric,cave2020ai,cave_craig_dihal_dillon_montgomery_singler_taylor_2018,dillon2020eliza,dillon2023ai,Bareis_2022_TalkingAI,guenduez2023strategically,vannoort2024use,chuan2019framing,robertson2023ai,taylor_automation_2018}.} This has been understood with regard to chatbots since the very first one, Joseph Weizenbaum’s ELIZA \cite{weizenbaum1966eliza,weizenbaum_computer_1987,dillon2020eliza,stone2024origins}. In 1995, Douglas Hofstadter \cite[p. 157]{hofstadter_fluid_1995} coined the term ``the Eliza effect,'' to name ``the susceptibility of people to read far more understanding than is warranted into strings of symbols – especially words – strung together by computers.'' The Eliza effect is in full swing around LLM-based chatbots, encouraged by the DMS-paratext. For example, consider the use of the first person in responses, which is only a convention of the DMS, and could easily have been implemented differently. The Eliza effect is also encouraged by other paratextual elements such as the hype from companies, researchers, and elite cuers, and the language of media coverage, much of which frames LLM-based chatbots as if they possess intention, knowledge and reasoning capabilities.\footnote{For just one recent example of the kind of language that encourages the Eliza effect, consider OpenAI’s \cite{OpenAI_o1_2024} statements on the \textit{reasoning} and \textit{thinking} capabilities of their o1 model.}  

Consider, for instance, the idea of ``hallucinations,'' a term widely used to designate the factual errors of LLM-based chatbots.\footnote{For examples of hallucinations see \cite{weiser_heres_2023} and \cite{alkaissi2023artificial}. For an example use of the term in technical contexts see \cite{openai2023gpt4}; for its use in the media, see \cite{weise_when_2023}.} As with other anthropomorphic terms to describe machines, the term ``hallucination'' reinforces the Eliza effect.\footnote{Consider for example the history and rhetoric of the idea of machine’s ``learning,'' as discussed in \cite[p. 4]{dillon2020eliza}.} A ``hallucination,'' as specified in the \textit{OED}, is a ``mental condition of being deceived or mistaken, or of entertaining unfounded notions'': to hallucinate, one must have a mind and notions. The LLM has neither, so although the DMS-paratext might emulate the language-games associated with them, the invocation of a mental life is simply more bullshit.

\section{A Masterman Semantic Detector}

We have dissected the anatomy of the LLM-based chatbot into two parts: the generalised model trained on many kinds of language-game, and the DMS-paratext that maintains the Eliza effect by emphasising a particular kind of game. We suggest that this is the game of bullshit, slop, ``sycophancy,'' and perhaps also of ``hallucination.'' The next part of our project explores whether it is possible to construct a semantic detector, of the kind proposed by Margaret Masterman, to identify when this language-game is being played. In honour of her contributions, we describe this technically as a Masterman Semantic Detector (MSD), although we also propose a rhetorical application of the method as a ``BS-meter.'' 

We use the corpus analysis methods pioneered by Masterman and Spärck Jones to characterise the ways that particular \textit{words} are used in particular social \textit{contexts}, with two kinds of machine learning model - one that identifies characteristic word frequencies, and a second that identifies characteristic contexts in which words are used. In order to train these models, we created a controlled corpus of ChatGPT output, constructed specifically to encourage the language-game of bullshit, taking Frankfurt’s \cite[p. 63]{frankfurt2005bullshit} definition of the bullshitter as a person obliged to ``talk without knowing what he is talking about.''

It may be surprising that we use \textit{AI output} to study \textit{human behaviour}. We compare this to a strategy that has proven effective for recognising patterns in human body movements (physical gestures), but in our case applying this strategy to recognising patterns in language. In the development of products such as Microsoft’s Kinect, digital simulations of human behaviour are used to generate synthetic or enriched datasets, larger than could economically be collected from real people, for use as simulated examples to train a neural network. In the case of Kinect, training sets of human movement were created, not by filming real humans, but automatically creating CGI animations of humans performing characteristic actions \cite{shotton2011real}. The patterns of movement that are familiar to any player of an Xbox videogame, such as kicking a ball or swinging a bat, can be reliably recognised because so many variants of the component motions were artificially simulated in order to train the recogniser. In adapting this strategy to language, we train a neural network with many \textit{simulated} examples of the semantic categories we are interested in. Whereas the Microsoft Kinect team used CGI to animate simulated human bodies, we use ChatGPT as a tool to ``animate" simulated human writing. This strategy of using a linguistic artefact (ChatGPT) as an instrument to model human language can be compared to Masterman's earlier use of an artefact (the English thesaurus) to model semantic content.

\subsection{The training set}

In contrast to the DMS-paratext that is designed to keep talking, veering from sycophancy and platitudes to apologetic contrition when it makes errors or has no facts to contribute \cite{sharma2023towards,malmqvist2024sycophancy,OpenAI_sycophancy_2025}, we looked for an exemplar of a language-game that is precise, factual, clear and concise – the opposite of the bullshitters described by Frankfurt. As an exemplar of precision, clarity and concision, that also undeniably reports (newly observed) scientific facts, we have chosen a basis for comparison in the house style of \textit{Nature} magazine – whose strict editorial standards make it possibly the most widely read and respected scientific journal in the world.

We constructed the first half of our training dataset by collecting 1,000 articles published in \textit{Nature}. To compare the DMS-paratext to this language-game of prestigious international science, we then prompted the latest version of ChatGPT (at the time of our experiment, the 4o release) to write an article for \textit{Nature} magazine, with the same title as an actual article amongst the 1,000 we had selected. To ensure that this emulated the style of a \textit{Nature} article, we included \textit{Nature}’s instructions to authors in the prompt. The resulting text, even to someone with minimal scientific training, was obviously bullshit. It included tables of fabricated data observations, some outright lies, unconnected arguments, but also long passages of vague but ``science-y'' text typical of the writing of weaker students. Most of these features have become familiar to those using these products (or whose students use them). Taking together the \textit{Nature} articles and ChatGPT’s pseudo-scientific fabrications, we obtained a dataset large enough to function as a training sample for state-of-the-art machine learning methods.

\subsection{The classifier}

XGBoost, the first algorithm we trained for our experiments, is widely used in machine learning for simple text classification. For these tasks, it primarily relies on TF*IDF (``term frequency–inverse document frequency''), a measure of the extent to which a particular word is distinctive of the document in which it appears. TF*IDF is attributed to Masterman’s student Spärck Jones \cite{sparck1972statistical}, now celebrated as a pioneering computer scientist, and famous for her observation that ``[c]omputing is too important to be left to men'' \cite{sparckjones2007computing}. It is no accident that the earliest linguistic critique of computational language models reported by Emily Bender, Timnit Gebru and colleagues in their \textit{Stochastic Parrots} critique \cite{bender2021dangers} is a paper by Spärck Jones \cite{jones2004language}, raising questions similar to our own, and also in the spirit of Masterman, with regard to the semantic content of generative models.

We thus use XGBoost to create a statistical model of the terms that best distinguished genuine \textit{Nature} articles from those fabricated by ChatGPT. We followed standard practice by removing uninteresting ``stop-words'' such as ``and,'' ``the,'' etc. from the TF*IDF dictionary. We also removed a small number of words that simply reflected the formatting of \textit{Nature} articles. (In particular, we found that our earliest iteration of XGBoost could identify \textit{Nature} articles with high confidence because they always included the word ``figure'' or ``fig,'' often followed by an alphanumeric sequence such as ``1a,'' ``2b,'' etc. The ChatGPT output, being plain text without figures, did not use these words.) The resulting classifier is 100\% accurate in cross-validation of held-out samples, and reports high confidence (99.84\%) in judging further examples constructed the same way.

Our second classifier, a fine-tuned RoBERTa transformer model \cite{liu2019roberta}, instead determines whether a text is closer to \textit{Nature} articles or ChatGPT output by analysing the structural context of word use rather than the words themselves.\footnote{We suggest that this innovative capability of the transformer architecture reflects Masterman’s own concern for how it might be possible to represent context in computation (e.g. \cite[p. 124]{masterman2005language}).} The RoBERTa transformer architecture, derived from the same original technical advances as today’s LLMs, processes text by encoding each ``token'' (a word or subword unit) as a multidimensional vector that captures the surrounding context of other tokens. By classifying a text based on the similarity of its contextual embeddings to those found in \textit{Nature} papers and ChatGPT outputs, RoBERTa effectively compares the structural properties of those linguistic contexts. As with the word-frequency-based classifier, the RoBERTa-based classifier is 100\% accurate in cross-validation, and also reports high confidence (99.97\%) in judging further examples.

\subsection{The MSD metric}

\begin{figure}[h]
  \centering
  \includegraphics[width=1\textwidth]{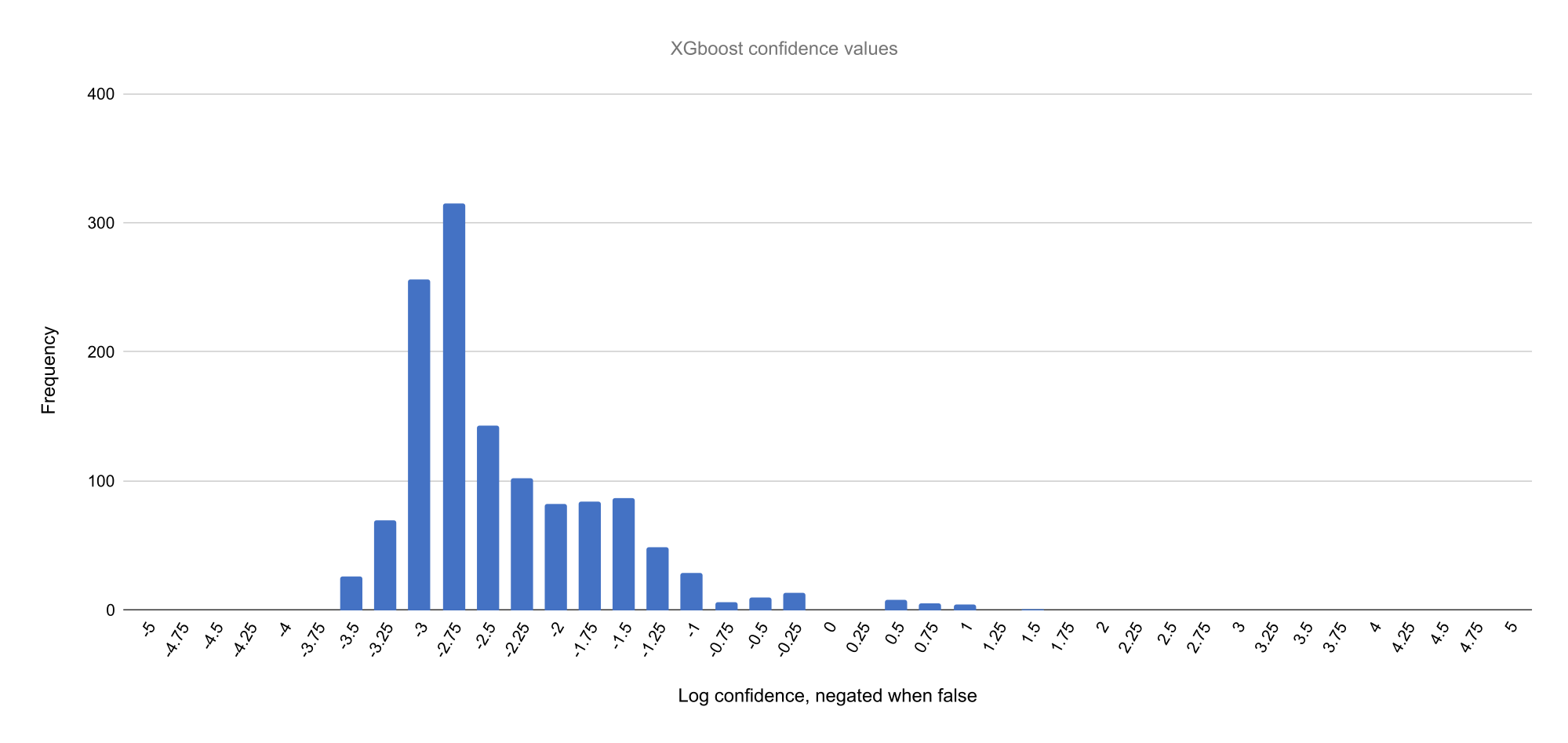}
  \caption{Frequency distribution of (log transformed) confidence scores for word classifier}
  \label{fig:word-classifier-frequencies}
\end{figure}

\begin{figure}[h]
  \centering
  \includegraphics[width=1\textwidth]{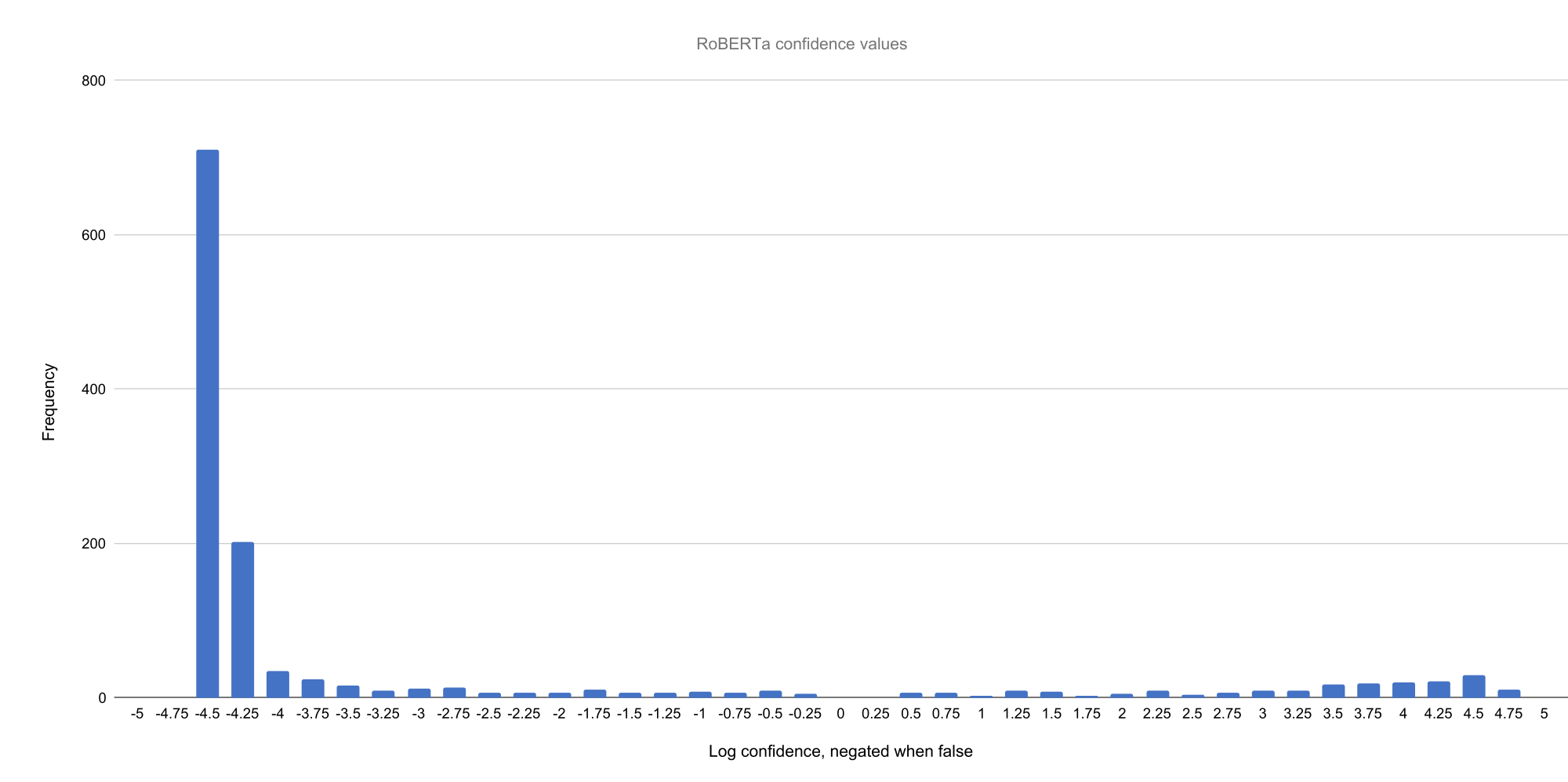}
  \caption{Frequency distribution of (log transformed) confidence scores for contextual classifier}
  \label{fig:context-classifier-frequencies}
\end{figure}

\begin{figure}[h]
  \centering
  \includegraphics[width=1\textwidth]{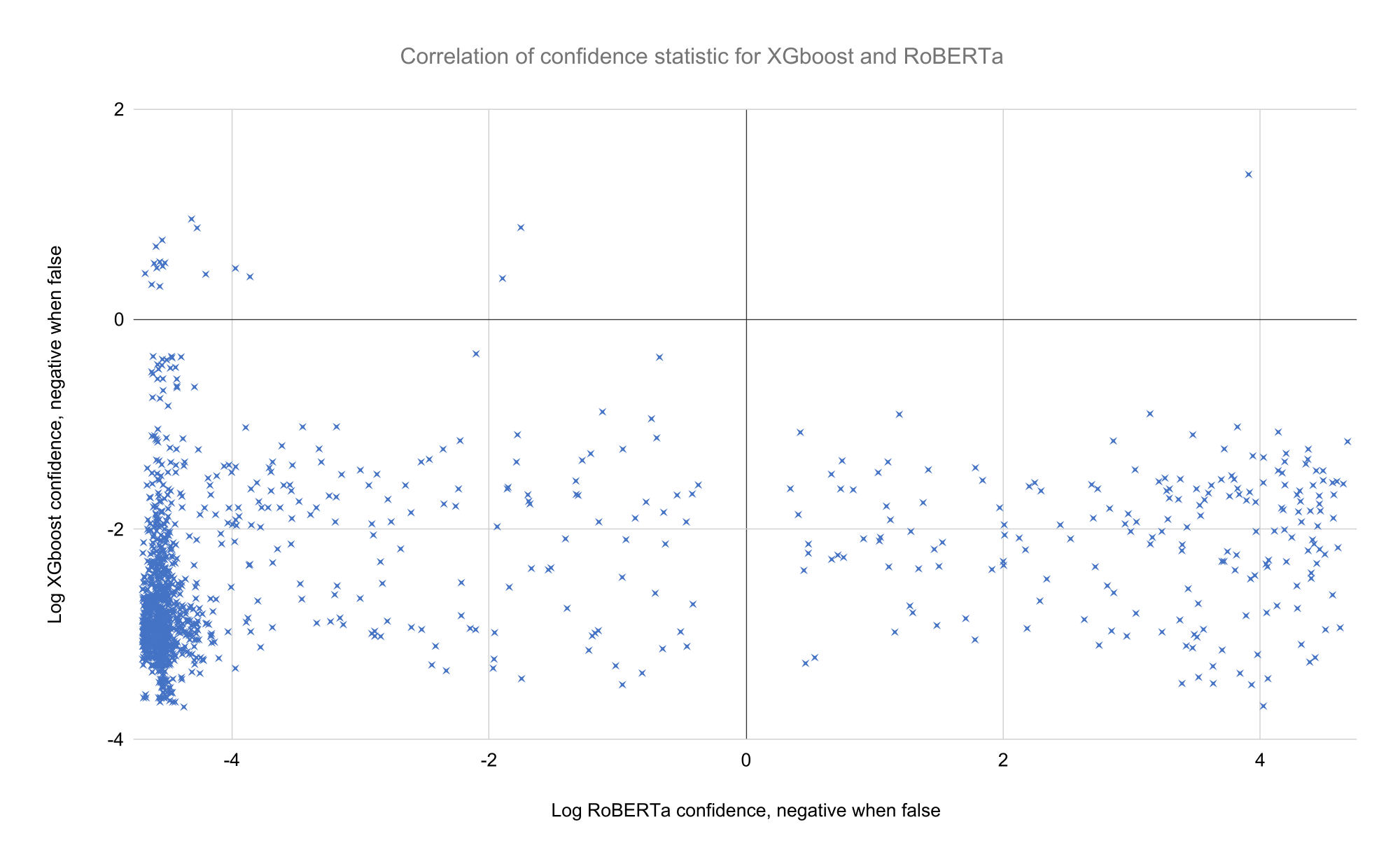}
  \caption{Correlation of (log transformed) confidence scores for word and context classifiers}
  \label{fig:scatter-word-context}
\end{figure}

Fig. \ref{fig:word-classifier-frequencies} shows the distribution of (log transformed) confidence scores from the XGBoost classifier,  for an exploratory experimental corpus containing many of the texts discussed in this paper. Fig. \ref{fig:context-classifier-frequencies} shows the distribution of confidence scores from the RoBERTa classifier. As seen in the scatter plot of Fig. \ref{fig:scatter-word-context}, there is only a small correlation between the confidence values of the two different classifiers (r = 0.282), meaning that they are basing their judgement on different language features (word frequencies in one case, and token embeddings in the other). Our MSD therefore combines the outputs of these independent elements.

Whereas many machine learning classifiers report only the most likely output class (if it passes some confidence threshold) rather than the value of the confidence statistic, our interest in the implicit encoding of language-games means that we are interested in the confidence values themselves. These can be taken as a computed information-theoretic summary reflecting the internal weights and activations within the models. The graphs in Figs. \ref{fig:word-classifier-frequencies}, \ref{fig:context-classifier-frequencies} and \ref{fig:scatter-word-context} summarise the values observed for this information measure on a scale from -5 to 5, where the midpoint of 0 would imply that the test item has exactly equal similarity to both the ChatGPT training set and to \textit{Nature} articles. Values above 0 are the log of the confidence statistic for a classification that the text is more like the ChatGPT training set, while values below 0 are (negatively) proportional to the log of the confidence if the text is classified as more like \textit{Nature} articles.\footnote{For reference, the formula used to calculate this score is as follows (note that, because confidence values are expressed as a probability between 0 and 1, the log of that value is a negative number): IF (class == ``bullshit'') THEN score = offset - scale * LOG(1-confidence) ELSE score = offset + scale * LOG(1-confidence).} Later in this paper, we present this value rhetorically as a ``BS-meter,'' with a percentage value better suited to public discussion, by simple linear conversion of the [-5, 5] range to [0, 100]. In future, a more powerful discriminator could be created using optimisation approaches rather than this simple linear statistic, but our results in the following experiments show that even a simple average of these two values can be used as a surprisingly accurate MSD.

To be clear, we are not stating that LLM technology can only ever simulate the language-game of bullshit. The LLMs encode all kinds of language-games, and, in principle, different kinds of DMS could select or emphasise these. However, the DMS-paratext in the present generation of LLM-based chatbots does prioritise the language-games of bullshit - and this is what our MSD detects. We now investigate whether this BS-meter can be used to detect cognate language-games — not only in LLM-based chatbot output, but in contexts where humans themselves may speak in this way.

\section{Experiment 1: The Language of Politics}

Now that our MSD approach has been defined, the question is whether we can detect the same language-game in other contexts. To do so, we apply the MSD to the classification of new texts. In our first experiment, we theorise and set out to test empirically a degree of affinity between the language used by ChatGPT and that used in political campaigning. In choosing the language of politics, we build on the work of George Orwell, ``perhaps the major contemporary forerunner of an approach to bullshit that focuses on the text itself'' \cite[p. 247]{fredal2011rhetoric}. Note that, for rhetorical purposes, we use the BS-meter values in numerical reports of the MSD analysis.

In Orwell’s writings, and especially in ``Politics and the English Language'' \cite{orwell_collected_1969v4}, political speech is described as an intentionally uninformative form of language. This proposition will hardly come as a surprise: Frankfurt himself \cite[p. 22]{frankfurt2005bullshit} suggests that bullshit is often found in ``advertising, public speech, and the nowadays closely related realm of politics.''\footnote{As one anonymous reviewer noted, political discourse and the behaviour of politicians may have shifted since Orwell’s critique, insofar as contemporary politicians are often characterised as engaging in outright lying rather than bullshitting. However, for the purposes of this paper, we focus on that more “traditional” conception of political discourse, which we take to be a more immediate point of comparison for LLM-generated language.} 

\subsection{Comparing political party manifestos to everyday spoken English}

As an experimental sample, we selected from a corpus of written English reflecting one of the types of political discourse to which Orwell most objected – the manifestos published by political parties in UK general elections.\footnote{As examples of bad political writing, Orwell cites ``pamphlets, leading articles, manifestos, White Papers and the speeches of Under-Secretaries'' \cite[p. 135]{orwell_collected_1969v4}.} We obtained 45 party manifestos, spanning the years 1945 to 2005, from the Manifesto Project Database,\footnote{See \url{https://manifesto-project.wzb.eu} (thanks to Mark Gotham for suggesting this source). We used the Manifesto Corpus version 2024-1 \cite{lehmann2024manifestocorpus}, the Comparative Electronic Manifestos Project (CEMP) \cite{pennings2006cem} and the Manifesto Data Collection version 2024a \cite{lehmann2024manifestodatacollection}. We removed non-prose formatting such as numbers and bullet points.} and calculated the MSD score for each of them.

As a contrast to this corpus of political speech, we calculated the MSD scores for transcripts of spoken English as found in the British National Corpus (BNC).\footnote{See \url{http://www.natcorp.ox.ac.uk/}.} We selected the BNC to represent what Orwell called ``demotic speech'' \cite[p. 135]{orwell_collected_1969v3}, the speech of the ``average man'' (p. 135), of the ``workingman'' (p. 136). According to Orwell, everyday demotic speech is generally characterised by transparent and communicative uses of language, which politicians should strive to adopt in order to more effectively communicate with their citizens. If one suspects that the purpose of political speech is not to inform but to deceive,\footnote{As satirised by Orwell in the Newspeak of \textit{1984}.} then the conversations recorded in the BNC might be more genuine, produced for the primary purpose of communicating information. Many of them are transcriptions of school lessons, university tutorials and lectures, hobbyist or professional training sessions, and of casual conversations between family, friends, colleagues, and strangers. After excluding transcripts of political meetings (mainly local government and trade union meetings) and news reports (frequently involving political commentary), we randomly selected 45 BNC texts of similar length to the corpus of 45 party manifestos. Using the method described above, we calculated the MSD scores for each of these 45 texts.

As far as we know, none of these 90 texts were written or spoken by research scientists, and none were written by ChatGPT because they predate it. Although the MSD classifiers were trained only to recognise scientific text and ChatGPT-generated text, our hypothesis is that the traces of the language-game selected by the DMS-paratext of the LLM-based chatbot to construct factually ungrounded ``bullshit'' science may resemble the uses of the English language in political discourse that were criticised by Orwell, and that he contrasted with everyday speech. Our experiment therefore tests the hypothesis that these two kinds of language-game may have statistically differentiable features (words and their context) that we can generically consider as bullshit in the Frankfurtian sense, whose traces can be measured through the application of the MSD method.

The null hypothesis (H0) is that the BS-meter scores will be the same for non-scientific, non-ChatGPT text regardless of whether the text comes from a political source or from a speech source. The alternative hypothesis (H1) is that the BS-meter scores will be different in samples of political texts, by comparison to samples of everyday speech.

\subsection{Results}

\begin{figure}[h]
  \centering
  \includegraphics[width=0.8\textwidth]{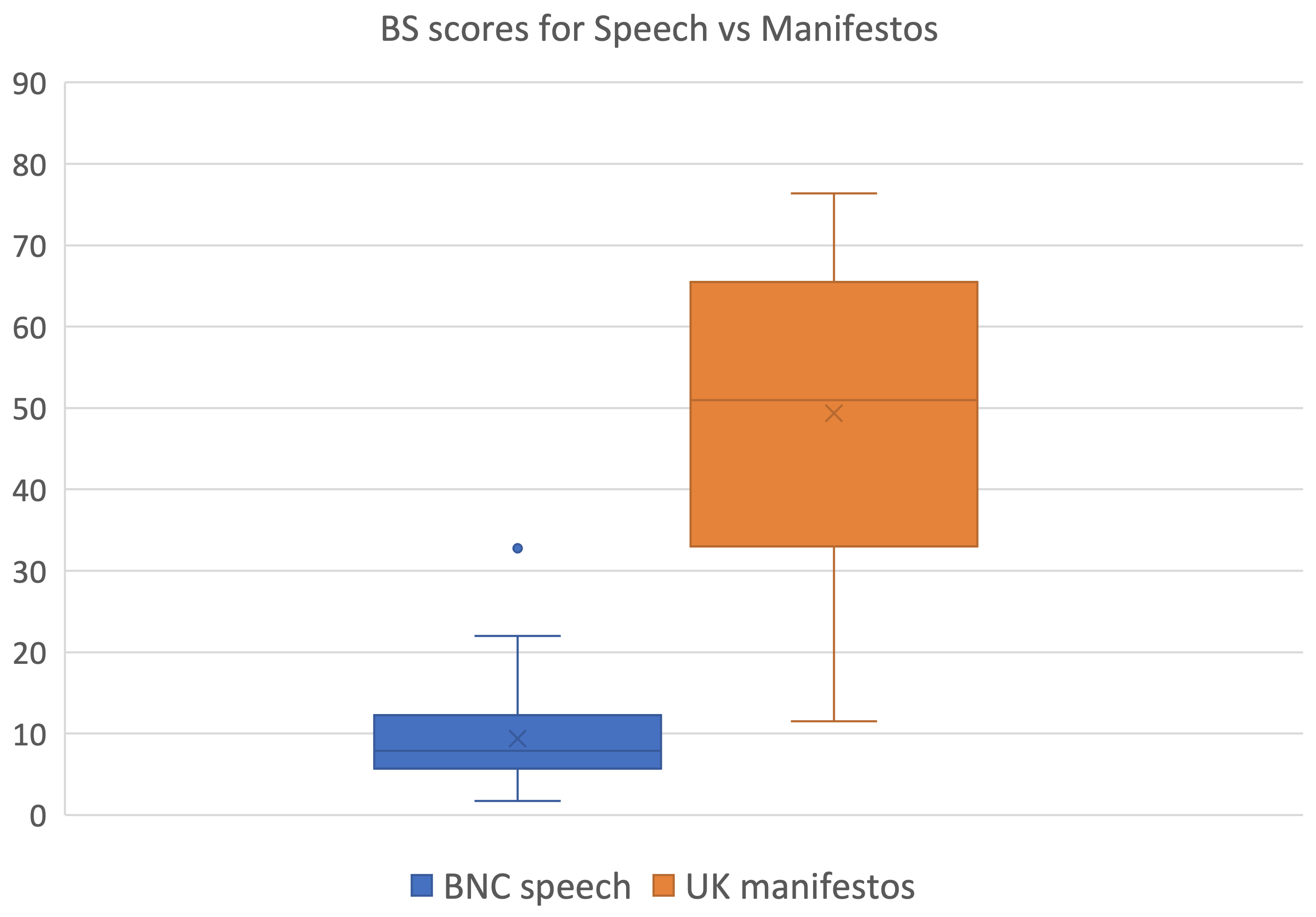}
  \caption{Comparison of BS-meter scores calculated for 45 samples of everyday UK speech from the British National Corpus, and 45 manifestos published by UK political parties.}
  \label{fig:orwell-bs-scores}
\end{figure}

As shown in Fig. \ref{fig:orwell-bs-scores}, the average BS-meter score for UK political manifestos is 49.36, while the average BS-meter score for non-political speech data from the British National Corpus is far lower, at 9.40. This difference is highly significant (t(54)=18.18, $p \ll 0.001$), meaning that we can reject the null hypothesis. 

We conclude, therefore, that the uses of the English language criticised by Orwell do share statistical properties with our training dataset. We state this cautiously, as appropriate to scientific investigation, and without wishing to suggest that correlation reflects causation. Nevertheless, the remarkably high statistical significance of our experimental finding, and the large difference in means on our BS-meter scale of 0-100, do indicate that the method is reliably able to capture some property of language. Furthermore, it seems that this property, whatever we call it, is remarkably shared between the DMS-paratext of the latest LLM-based chatbots on one hand and, on the other, the uses of the English language criticised by Orwell more than 50 years ago. 

To emphasise how surprising this is, we note that one might expect the formal register, and tertiary education, of leading scientists to have more in common with the British political class, while the conversational style of ChatGPT might be expected to more closely resemble everyday speech. The opposite is true. Ordinary people speak like scientists, while politicians talk like ChatGPT.

\section{Experiment 2: Bullshit Jobs}

Our introduction presented the socio-linguistic-technical ``body'' of ChatGPT as having both political and economic significance. Our first experiment made a comparison to political use of language. This second experiment turns to the economic context in which language is used.

In \textit{Bullshit Jobs} \cite[pp. 9-10]{graeber_bullshit_2018}, David Graeber defines bullshit jobs as ``paid employment that is so completely pointless, unnecessary, or pernicious that even the employee cannot justify its existence even though, as part of the conditions of employment, the employee feels obliged to pretend that this is not the case.'' Shannon Vallor's \cite[p. 121]{vallor2024ai} comparison between the rhetorical style of LLM-based chatbots and that of a smooth and oily car salesman is a useful starting point to investigate a possible parallel between the bullshit text resulting from the DMS-paratext of chatbots and the text created in bullshit jobs identified by Graeber. A car salesman is indeed bullshitting according to Frankfurt, if he misrepresents his intentions and state of mind, appears falsely invested in a client’s interests and inflates the worth of the cars he sells. His job would also be classified as bullshit according to Graeber if it does not contribute any social value. For example, the job itself might complicate a process that could be more straightforward if the client could just get a precise, clear and concise description of the condition and characteristics of a car. Like the car salesman imagined by Vallor, the DMS-paratext of LLM-based chatbots has been designed to produce text that persuades customers as to its competence, adjusted to interact with users amiably and with a veneer of sycophancy or professionalism but not, crucially, to be truthful. Just like someone in a bullshit job, the bullshit-generating DMS-paratext added to the LLM is designed to produce an illusion of meaningful work.

We designed a second hypothesis-testing experiment, to compare the Frankfurtian language of bullshit to the text resulting from the Graeberian employment conditions of the bullshit job. Our hypothesis is that text produced by those employed in bullshit jobs is itself more likely to have the semantic language-game characteristics of bullshit text. As an objective measure of this MSD we again use the BS-meter scores described above.

\subsection{Measuring the language-games played in bullshit jobs}

We collected 100 sample texts from online sources, 50 of which were selected as likely to have been written by people employed in bullshit jobs (as characterised by Graeber). A control sample of 50 further texts were selected as likely to have been written by people employed in professions that would not fall within the scope Graeber defines as bullshit. None of the 100 texts were written by scientists, and none of the 100 texts (so far as we were aware, given the challenges of precise provenance and dating for informal online publications) were written by ChatGPT.\footnote{However, note that, as discussed below, at least one of our control texts seemed as though it might have been created in part using an LLM.} We used this sample of texts from 50 hypothetically bullshit jobs and 50 hypothetically non-bullshit jobs to test the hypothesis that these two samples have statistically differentiable language features.

Identifying the experimental sample of texts required considerable research judgement. Graeber’s book describes the characteristics and status of bullshit jobs at substantial length, including individual case studies as well as survey findings and interview research. In one section of the book, he offers a typology of five particularly frequent classes of bullshit job. However, this typology is neither strictly defined, nor necessarily rigorous, and certainly not presented in a way designed for investigation through hypothesis-testing experimentation. We therefore used the more succinct summary version of Graeber’s typology, developed through consensus by the editors of the Wikipedia article ``Bullshit jobs,'' to serve as the working definition for our sample construction (the category labels are Graeber’s own):

\begin{itemize}
 \item \textit{Flunkies}, who serve to make their superiors feel important, e.g., receptionists, administrative assistants, door attendants, store greeters;
 \item \textit{Goons}, who act to harm or deceive others on behalf of their employer, or to prevent other goons from doing so, e.g., lobbyists, corporate lawyers, telemarketers, public relations specialists;
 \item \textit{Duct tapers}, who temporarily fix problems that could be fixed permanently, e.g., programmers repairing shoddy code, airline desk staff who calm passengers with lost luggage;
 \item \textit{Box tickers}, who create the appearance that something useful is being done when it is not, e.g., survey administrators, in-house magazine journalists, corporate compliance officers;
 \item \textit{Taskmasters}, who create extra work for those who do not need it, e.g., middle management, leadership professionals.\footnote{See \url{https://en.wikipedia.org/wiki/Bullshit\_Jobs}, accessed 22 November 2024.} 
\end{itemize}

Although Graeber did not pay substantial attention to the other kinds of work in society that constitute \textit{non-bullshit} jobs, these are mentioned from time to time in his text. It is also possible to propose general principles from the social sciences that are likely to be consistent with Graeber’s critical orientation, for example professions that directly deliver services at the base of Maslow’s hierarchy of needs, or professions where the employee is directly engaged in useful labour rather than in supervising or managing others.

For each of these five classes, we therefore selected 10 texts that we would expect to have been written by or on behalf of those employed in that class, and 10 contrasting texts that demonstrated the opposing, non-bullshit principles. There is some danger that our selection process might be considered as derogatory or libellous, since the original author can often be directly identified. For that reason, we are not publishing this experimental dataset, but will be happy to make it available on request, for replication purposes. The specific rubrics that we used to collect these 10 sets of sample texts are reported in Table \ref{tab:jobs}.

 \begin{table}[h]
  \centering
  \renewcommand{\arraystretch}{2}
  \begin{tabular}{|l|p{5cm}|p{5cm}|}
    \hline
    \textbf & \textbf{Bullshit sample} & \textbf{Contrast sample} \\
    \hline
    Flunkies & We searched for corporate biographies of prominent chief executives, hypothesising that these are likely to have been drafted by a flunky expected to make the subject feel important. & We searched for historical biographies of engineers, avoiding those who had founded companies, were famous household names, or celebrated for reasons other than their practical contributions. \\
    \hline
    Goons & We searched for examples of corporate mission statements, hypothesising that these have been written by some combination of lobbyists, lawyers, and PR specialists. & We selected text from organisations whose purpose is to directly report the truth against opposition, including whistleblowers, public health educators, and human rights information services. \\
    \hline
    Duct tapers & We searched for policies for temporary repair and delays, management of complaints, refunds, compensation, and samples of how to write business apologies. & We searched for practical instructions on how to directly fix problems with appliances, construct things, or achieve immediately useful results. \\
    \hline
    Box tickers & We selected text written by or about corporate compliance officers who had been recognised with prizes in the International Compliance Association awards. & We sampled writing by people who create goods at the base level of Maslow’s hierarchy of needs including farmers, agricultural and food processing engineers, chefs, water treatment engineers, builders, garment manufacturers and nurses. \\
    \hline
    Taskmasters & We selected personal statements or blog entries from the company websites of award winners in ``Global Gurus Top 30 - the World's Top 30 Leadership Professionals.'' & We sampled job descriptions that do not directly include leadership, delegation or management, including building construction, food preparation, laundry, and road maintenance. \\
    \hline
  \end{tabular}
  \vspace{1em}
  \caption{Sample rubrics used to collect texts associated with bullshit jobs}
  \label{tab:jobs}
\end{table}

Although the text sampling procedure involved a significant element of research judgement in applying and interpreting Graeber’s analysis, there is no reason to expect that either of these two samples of text would be more or less \textit{Nature}-like, or more or less ChatGPT-like, or that either would have greater preponderance of any particular MSD features. Our sampling procedure did not involve any prior judgment on the basis of linguistic properties, or assume that there would be any statistically observable differences at all.

The null hypothesis (H0) is that the BS-meter scores will be the same for non-scientific, non-ChatGPT text regardless of whether the text is written by someone with a (Graeberian) bullshit job or a non-bullshit job. The alternative hypothesis (H1) is that the BS-meter scores will differ according to the class of jobs from which our textual samples are taken.

\subsection{Results}

We performed a repeated-measures analysis of variance (ANOVA), with two factors as independent variables. One independent variable was bullshit/non-bullshit contrast, with two values, and the other variable was the Graeber category, with five possible values. The dependent variable was the BS-meter score.

We observed a highly significant main effect (F(99,1)=43.73, $p \ll 0.001$), meaning that we can reject the null hypothesis with extremely high confidence. The effect size is large, with an overall mean BS-meter score of 52.47 for bullshit jobs, compared to 28.87 for the contrast sample. We can conclude that samples of text selected to represent employment in Graeberian bullshit jobs do resemble the Frankfurtian bullshit produced by ChatGPT’s DMS-paratext far more than they do precise, factual and clear scientific writing. Conversely, text produced by those in what we potentially identified as non-bullshit jobs, such as laundry, cleaning or road repairs, have more resemblance to top-quality scientific writing than to ChatGPT output.

\begin{figure}[h]
  \centering
  \includegraphics[width=0.8\textwidth]{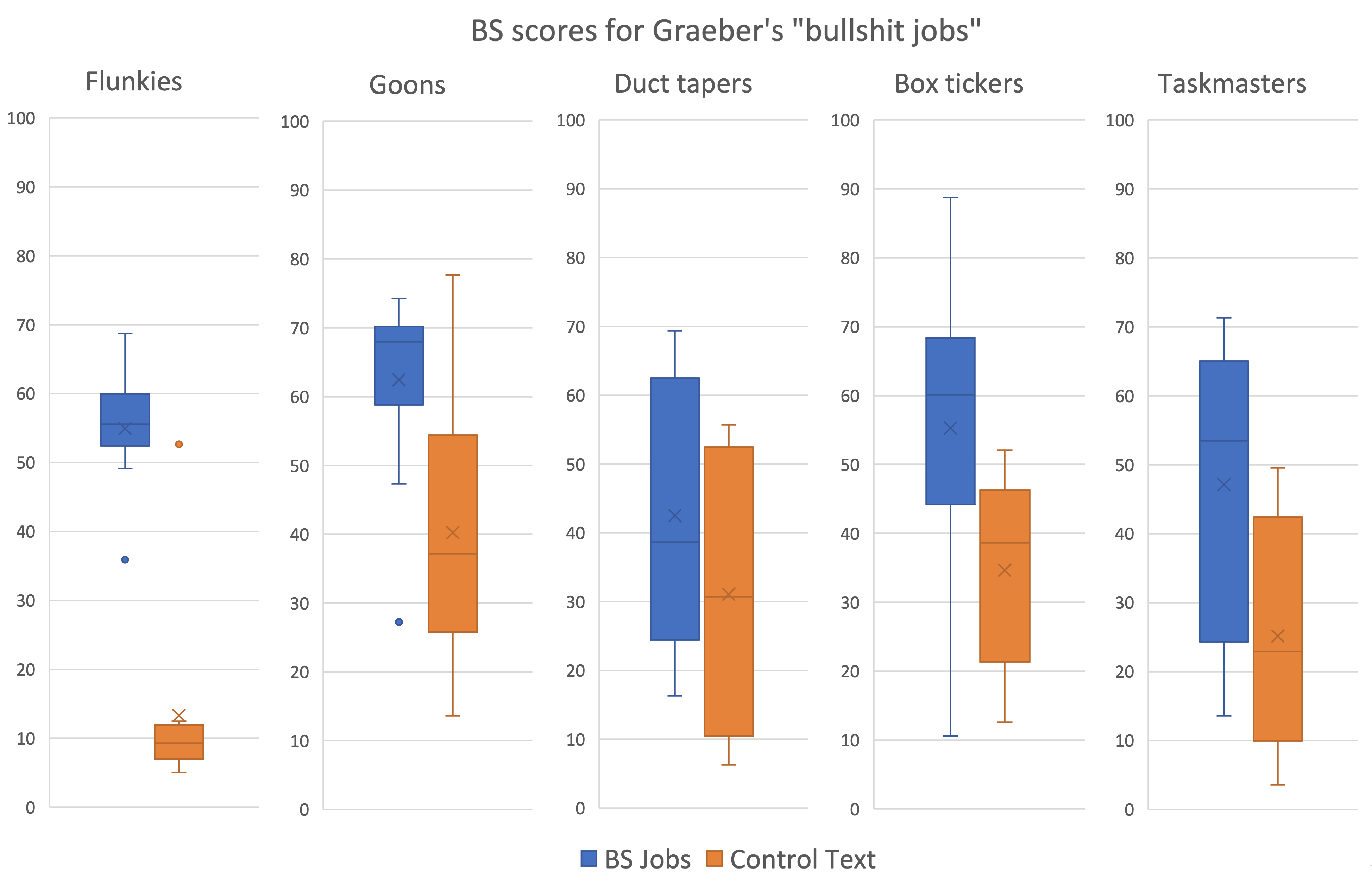}
  \caption{Comparison of BS-meter scores for text produced in the five classes of Graeber’s bullshit jobs, as compared to contrast texts in other jobs.}
  \label{fig:graeber-bs-scores}
\end{figure} 

We also observed a weaker effect of differences in BS-meter scores over the five Graeber classes (F(99,4)=3.30, p=0.014), which interacts with marginal significance with the strong bullshit effect (F(99,4)=1.92, p=0.114). As seen in Fig. \ref{fig:graeber-bs-scores}, this interaction results in large part from the variability in scores for the selected non-bullshit control texts. In all five categories, there are large differences in means between the hypothetically bullshit texts and the selected comparison texts. In order to investigate these further, we carried out a post-hoc t-test to compare the means of the 5 groups of 20 contrasting samples, as shown in Table \ref{tab:scores}.

\begin{table}
  \centering
  \renewcommand{\arraystretch}{1.5} 
  \begin{tabular}{|l|c|c|c|}
    \hline
    \textbf & \textbf{Mean BS score} & \textbf{Mean contrast score} & \textbf{Significance} \\
    \hline
    Flunkies & 54.96 & 13.28 & $p \ll 0.001$ \\
    \hline
    Goons & 62.43 & 40.22 & $p < 0.05$ \\
    \hline
    Duct tapers & 42.50 & 31.08 & $p = 0.23$ \\
    \hline
    Box tickers & 24.38 & 20.71 & $p < 0.05$ \\
    \hline
    Taskmasters & 47.16 & 25.14 & $p < 0.05$ \\
    \hline
  \end{tabular}
  \vspace{0.5em} 
  \caption{Post hoc comparison of BS-meter scores for each of the five Graeber classes of bullshit jobs. p-values under 0.05 (5\%) are considered statistically significant. A p-value far less than 0.001 is highly significant.}
  \label{tab:scores}
\end{table}

We report the varying significance scores between the five classes for completeness. The significance values in Table 5 might possibly be taken as evidence that, while every one of the five Graeber classes seems to produce more bullshit on average, perhaps some text produced in the duct tape category is less sloppy than in other categories. However, the variability in our sample construction procedure (which could certainly be made more rigorous in future) might easily account for these differences. For example, the non-BS texts we selected for comparison to duct tapers turned out to include three texts with high BS-meter scores: advice on how to change your car’s oil that was published by a car insurance company; advice on how to build a wall that was published by a building materials supplier; and first aid advice that was published by a wealthy private health clinic. In each case, we might ask whether the incentive structures for these publishers (and the anonymous corporate-style advice they publish) makes them a relatively poor comparator. In fact, for the second of these, the AI detector GPTZero reports an 84\% probability that the wall-building text was generated by AI. Future investigations using the MSD method, especially if replicating our application of the method to Graeber’s studies of work, should take further care in selecting the texts used.

\section{Conclusion – A BS-meter?}

Many commentators have already observed that AI as recently manifested in LLM-based chatbots appears to produce bullshit, more politely describing such text as AI slop. Although this may be argued in relation to the probabilistic mechanism by which LLMs generate text, which is indifferent to truth or falsity—an indifference often apparent from even a cursory inspection of their outputs—our concern is not purely epistemic. We seek to investigate how the \textit{language-game} of bullshit comes into being in these systems and how it may best be characterised. In this paper, we have demonstrated how the statistical methods of natural language processing, heavily influenced by the language philosophy of Margaret Masterman, can be used as a semantic detector to study the language-game of how words are used in social contexts. We show how the word frequencies and embeddings encoded in LLMs can be used to detect traces of different human language-games. With regard to the specific language-game of bullshit, we explain how it is amplified and prioritised by the paratextual apparatus of the dialogue management system that presents LLM sequence prediction as a (supposedly) artificially intelligent chatbot. By asking ChatGPT to generate scientific articles about recent discoveries it does not know of, we are able to provide a reference set of how this bullshit is manifested. We then trained two different classifiers by contrasting that reference set of bullshit to a large collection of factual, precise, clear and concise scientific writing. 

We find that the resulting Masterman Semantic Detector can be applied as a remarkably reliable BS-meter. Although our reference set was constructed according to the rubric of Frankfurt, by requesting speech on a topic where the speaker has no knowledge, we can only say that the detector reliably detects some patterns of language use, not necessarily what language-game this reflects. However, our experimental investigations show, firstly, that the language-game detected by the BS-meter is reliably present in the political misuse of English castigated by Orwell and, secondly, that this language-game is more likely to be seen in professional writing by those people whom Graeber identifies as having bullshit jobs. That further coincidence, with its clear social relevance and significance in relation to future applications of LLM-based chatbots, offers compelling evidence that we really are experiencing and measuring bullshit.

\section*{Acknowledgements}
Harry Giddens’ contribution to this research was supported via the Google DeepMind Research Ready funding stream. Data used in Section 4 has been extracted from the British National Corpus Online service, managed by Oxford University Computing Services on behalf of the BNC Consortium. Following the completion of the first draft of this paper, an actual BS-meter was created by a team of students (Alex Newsham, Hayden Young, Martin Guenther, Nigel Arun Jacob, Wei Heng Wong), in the course of which they independently replicated our results with their own implementation of a classifier. We are grateful to Clayton Lewis for close reading and critique, and to anonymous reviewers for insightful discussion.

\bibliographystyle{apalike}
\bibliography{refs}

\end{document}